\documentclass[sigconf,screen,nonacm]{acmart}
\AtBeginDocument{%
  \providecommand\BibTeX{{%
    \normalfont B\kern-0.5em{\scshape i\kern-0.25em b}\kern-0.8em\TeX}}}
    
\usepackage{supertabular,booktabs}

\begin{document}

\title{Risk factor identification for incident heart failure using neural network distillation and variable selection}

\author{Yikuan Li}
\email{yikuan.li@wrh.ox.ac.uk}

\affiliation{
  \institution{Deep Medicine, University of Oxford}
  \country{}
}
\author{Shishir Rao}
\author{Mohammad Mamouei}
\author{Gholamreza Salimi-Khorshidi}
\author{Dexter Canoy}
\author{Abdelaali Hassaine}
\author{Thomas Lukasiewicz}
\author{Kazem Rahimi}

\affiliation{
  \institution{Deep Medicine, University of Oxford}
  \country{}
}

\renewcommand{\shortauthors}{Li, et al.}


\begin{abstract}
  Recent evidence shows that deep learning models trained on electronic health records from millions of patients can deliver substantially more accurate predictions of risk compared to their statistical counterparts. While this provides an important opportunity for improving clinical decision-making, the lack of interpretability is a major barrier to the incorporation of these ‘black-box’ models in routine care, limiting their trustworthiness and preventing further hypothesis-testing investigations. In this study, we propose two methods, namely, model distillation and variable selection, to untangle hidden patterns learned by an established deep learning model (BEHRT) for risk association identification. Due to the clinical importance and diversity of heart failure as a phenotype, it was used to showcase the merits of the proposed methods. A cohort with 788,880 (8.3\% incident heart failure) patients was considered for the study. Model distillation identified 598 and 379 diseases that were associated and dissociated with heart failure at the population level, respectively. While the associations were broadly consistent with prior knowledge, our method also highlighted several less appreciated links that are worth further investigation. In addition to these important population-level insights, we developed an approach to individual-level interpretation to take account of varying manifestation of heart failure in clinical practice. This was achieved through variable selection by detecting a minimal set of encounters that can maximally preserve the accuracy of prediction for individuals. Our proposed work provides a discovery-enabling tool to identify risk factors in both population and individual levels from a data-driven perspective. This helps to generate new hypotheses and guides further investigations on causal links.
  
\end{abstract}






\maketitle

\section{Introduction}
With the growing access to large-scale electronic health records (EHR) from millions of patients in recent years, deep learning provides an unprecedented opportunity to tackle risk prediction in a data-driven way~\cite{AYALASOLARES2020103337,gangwar2020deep,shickel2017deep}. Previous evidence demonstrated that deep learning models substantially outperform statistical models in terms of risk prediction~\cite{choi2016retain,Kwon_2019,kim2019cox,nguyen2016deepr,choi2016doctor,beaulieu2016semi}. Nevertheless, due to their deep, complex architecture, it is difficult to retrieve medical knowledge from these models for further clinical usage, hindering their wider applications in healthcare. Therefore, there is a need for further work on the interpretability of these models. 

Current model interpretation methods mainly focus on the discovery of a subset of relevant features and the quantification of their importance for a particular task. Some of the commonly used methods are sequential correlation feature selection~\cite{hall1999correlation}, mutual information feature selection~\cite{nagpal2018feature}, and Shapley values~\cite{vstrumbelj2014explaining}. However, these methods can only discover important features from a number of fixed, pre-selected features, thus, are not flexible enough to untangle patterns learned from longitudinal EHR with various number of visits and records for each patient. 

Heart failure (HF) is considered a major public health problem. Despite evidence indicating the decline in incidence and better survival rates, the prevalence of HF has been increasing~\cite{yang2015clinical,sahle2017risk,dunlay2014understanding}. Almost 920,000 people have been diagnosed with HF in the UK in 2018~\cite{factsheet} and the five-year mortality rate remained as high as approximately 50\%~\cite{butler2008incident,echouffo2015population}. Therefore, identifying early-stage risk for HF to guide decision making is important. Statistical models have been used extensively for the prediction of incident HF risk and the identification of associative and causal factors of HF. Such studies have contributed to a better understanding of some the determinants of HF~\cite{kalogeropoulos2010inflammatory,kannel1999profile,nambi2013troponin,agarwal2012prediction}. For instance, 27 clinical factors have been verified to be associated with incident HF in 15 different studies~\cite{yang2015clinical}. However, on average, their predictive performance has been unsatisfactory, outlining omitted variables or inadequate modelling strategies~\cite{sahle2017risk,echouffo2015population,levy2014heart}.

In this study, we proposed two methods, namely, model distillation and variable selection, to interpret a state-of-the-art probabilistic sequential deep learning model, BEHRT~\cite{li2020deep}, for HF risk association identification. To this end, BEHRT was trained on longitudinal linked EHR for the prediction of incident HF. Subsequently, we applied model distillation~\cite{hinton2015distilling,tang2019distilling} on BEHRT to extract the learned hidden patterns and incorporate them into a simpler, more interpretable model. In order to provide an uncertainty estimation for predictions as the probabilistic BEHRT model, we designed the model as a Bayesian deep learning model. The simple model was designed to combine the merits of two types of latent variables, namely, contextual variables and additive variables, for HF risk prediction. The contextual variable captured the complex temporal interactions of input features with the use of a bidirectional long short-term memory (BiLSTM)  neural network. The additive variable was constructed as the weighted sum of independent input features (with no interactions) where the weights provided an intuitive way to quantify the association of each feature to HF. The contextual variable and additive variable were then fed into a Gaussian processes (GP) classifier to produce a risk prediction that captured the two components. The motivation behind this design was to separate the additive elements of risk (as reflected by independent variables) from the elements that arise from more complex interactions between input features (captured by contextual variables). Before further investigation in interpretation and association identification, we validate the reliability of the model using area under receiver operating characteristic curve (AUROC) and area under precision-recall curve (AUPRC).

The identification of medical conditions that are associated or dissociated (i.e., negatively associated) with HF is carried out by analysing their contribution to the additive and the contextual variables.  For the additive variable, the contribution of each condition to risk of HF is captured by the corresponding coefficient in the weighted sum.  The contextual variable is a more abstract representation of nonlinear interactions between medical conditions over time, modelled by a BiLSTM. Therefore, we define the relative contextual variable risk of an event to HF as the ratio of the contextual variable in the event (e.g, diabetes, hypertension, or angina) exposure and non-exposure groups; similar to the notion of hazard ratio or relative risk in statistics. We mapped all diseases into an association map based on relative contextual variable and corresponding weight in additive variable (Section~\ref{sec:pop_association}). This map directly reflected the population-level association between a disease and incident HF. In addition, to further support individual-level decision making, we conducted a variable selection method to maximally preserve a patient’s temporal information while keeping as minimal critical records as possible. Therefore, with a much smaller number of records, the model can have a similar predictive performance as using the entire EHR. We present this as to give an overview of the structure of the paper.

\section{Methods}
\subsection{EHR dataset}
Our dataset was acquired from Clinical Practice Research Datalink (CPRD)~\cite{CPRD}. It includes records collected from general practices across the UK, and can be linked to hospital records and other secondary datasets such as the Office for National Statistics. To interpret the previously proposed BEHRT model for incident HF prediction, we followed the identical data pre-processing method in the original paper~\cite{li2020deep}.  In brief, we used records between 1985 and 2015, for both men and women who were at least 16 years old. Clinical information included diagnoses, medications, and date of birth. All the diagnoses were mapped to the 10th revision of the International Statistical Classification of Diseases and Related Health Problems (ICD-10)~\cite{bramer1988international} based on a previously published dictionary~\cite{li2020deep} and all medications were mapped to the British National Formulary (BNF) code in section level. Our dataset comprised 788,880 patients, 8.3\% of whom were diagnosed with incident HF during the 6-months prediction window (HF +). We randomly split the dataset into a 60\% training set, a 10\% tuning set, and a 30\% validation set.

\subsection{Model performance evaluation}\label{sec:evaluation}
The direct interpretation of a complex model is challenging. This work aims to transfer the predictive capability from an established complex model to a simpler and more interpretable model using knowledge distillation to identify the associations that underlie the model’s predictions. Therefore, the first concern would be the generalisability and predictive performance of the new model. In this analysis, we developed three models for comparison: an established probabilistic BEHRT as the reference model, a simple Bayesian deep learning model that distils knowledge from the BEHRT model, referred to as BDLD, and the same Bayesian deep learning model trained without distillation mechanism (referred to as BDL for simplicity)~\cite{tang2019distilling,korattikara2015bayesian}. The objective is to prove that with knowledge distillation, the simple BDLD model can achieve a comparable predictive performance and generalisability, therefore, it is a valid model that can be further used for the interpretation of risk associations. 

All models were trained and tuned using training and tuning set, respectively. Afterwards, we evaluated these models on the validation set and compared their predictive performance using AUROC, AUPRC, and the calibration curve. Since all probabilistic models generated a probability distribution for the risk of HF, the mean and the confidence interval were calculated as follows: (1) a random sample of size 30 was drawn from the predicted probability distribution of a given patient, and all metrics were evaluated using the mean probability over these 30 samples; (2) a random sample of size 1 was drawn from the predicted probability distribution of a given patient, all metrics were calculated based on this sample, afterwards, this process was calculated for 30 times, and we reported the mean and 95\% confidence interval over these 30 evaluations. The first method focused on evaluating a model’s average predictive performance, and the second method can show the stability of model predictions. Also, we did not see any material difference when we sampled more than 30 times.

\subsection{Model architectures}
This study involved two model architectures (3 scenarios): A Transformer-based BEHRT model, a Bayesian deep learning model trained without knowledge distillation (BDL), and the same Bayesian deep learning model trained with model distillation (BDLD). 

The BEHRT model is based on a recently proposed Transformer architecture~\cite{li2020deep}, which uses attention mechanism to extract long-term dependencies within sequential data. It treats a medical event as word, a visit as sentence, and the entire EHR as a document to mimic how natural language processing models model the sequential language. Instead of building sequential dependencies event by event like recurrent neural network~\cite{zia2019long}, it uses auxiliary features such as age, segmentation, positional code to represent the order, as well as the irregular interval between events. Additionally, the probabilistic BEHRT model uses stochastic embedding parameters and a Gaussian processes classifier for prediction. We refer readers to the original paper~\cite{li2020deep} for more details. 

\begin{figure}[h]
  \centering
  \includegraphics[width=\linewidth]{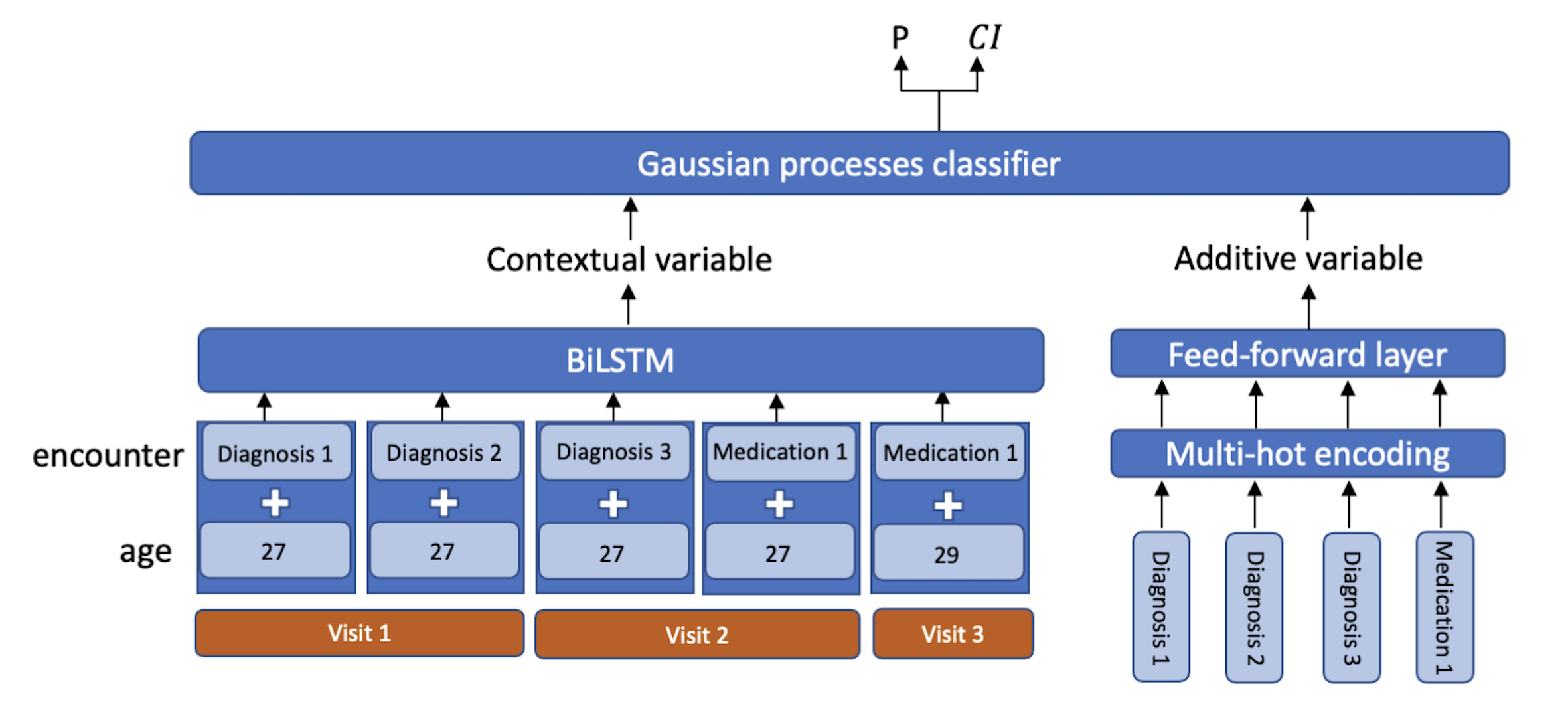}
  \caption{Bayesian deep learning model (BDL and BDLD) architecture. P represents the mean predictive probability; CI represents predictive 95\% confidence interval. For BiLSTM, we feed the events in the EHR as a sequence, with age to represent their order. For the feed-forward layer, we encode the medical events into a multi-hot vector, where each dimension represents whether an event occurred in a patient’s medical history. As shown, there are two medication 1 recorded in the patient’s history, however, for a multi-hot encoder, we only need to feed in if medication 1 is recorded or not, while the number of repetitions is irrelevant.}
  \label{fig:model_arch}
\end{figure}

For the BDL(D) model, as shown in Figure~\ref{fig:model_arch}, it included three components: a one-layer bidirectional long short-term memory network~\cite{zhang2015bidirectional}, a one-layer feed-forward network, and a Gaussian processes classifier~\cite{li2020deep,wilson2015deep}. The BiLSTM component, in which the embedding parameters are stochastic, was designed to extract the interactions of temporal features and summarise them as a contextual variable. The feed-forward network (weight parameters are stochastic) was equivalent to a linear component, which assigned a coefficient to each feature (predictor) to represent its association to prediction. Both risk scores are learned by the neural network. We summed the coefficient of events with occurrence in a patient’s medical records as a single score (additive variable) to represent risk learned from independent predictors. In the end, we used a Gaussian processes classifier to learn a multivariate correlation among the contextual variable, additive variable, and risk of incident HF.

\subsection{Model distillation}
\begin{figure}[h]
  \centering
  \includegraphics[width=\linewidth]{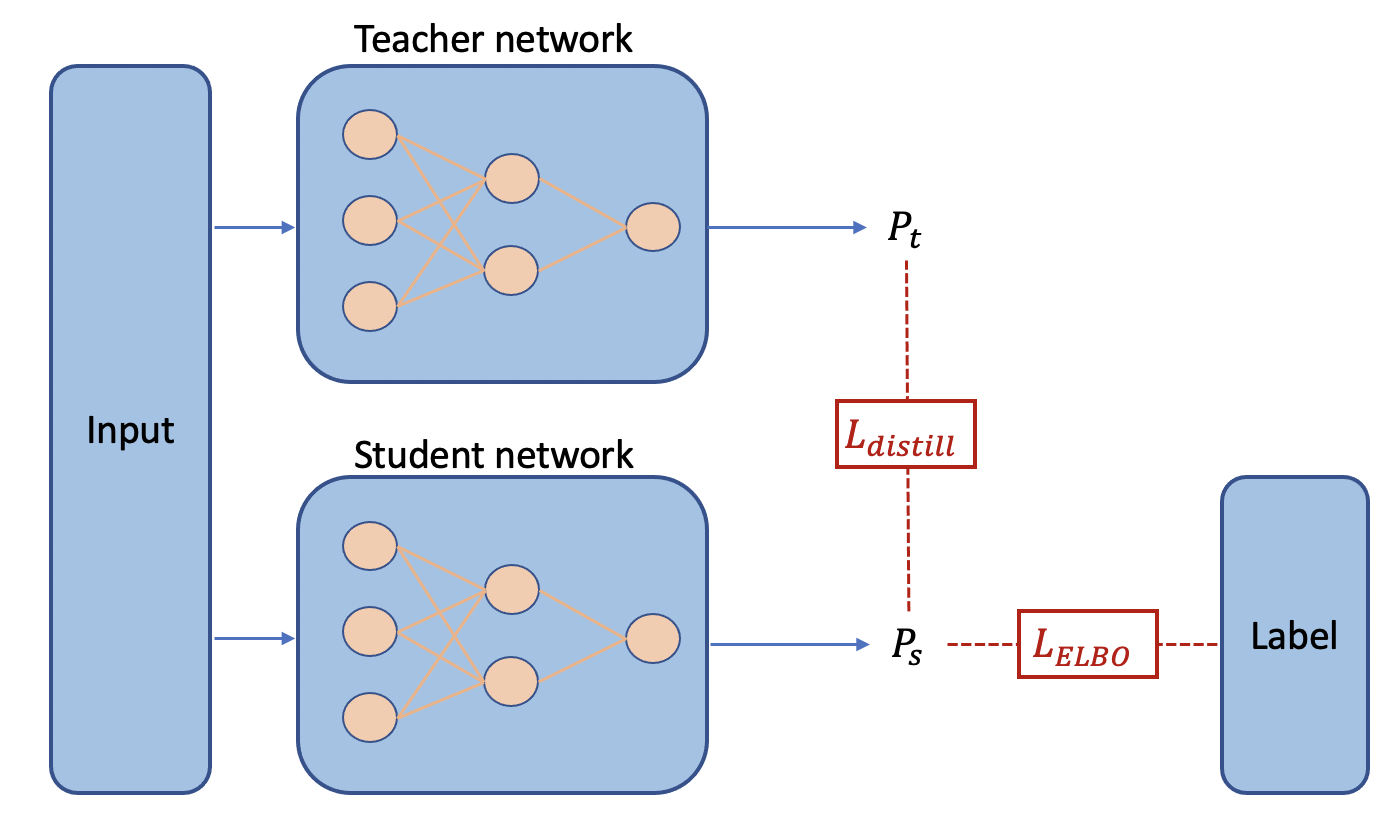}
  \caption{Illustration of model distillation. The figure shows the mechanism of model distillation. $P_{t}$ and $P_{s}$ represent a prediction sampled from the predictive distribution of the teacher network and the student network, respectively; $L_{distill}$ and $L_{ELBO}$ are two loss functions for distillation and evidence lower bound, respectively.}
  \label{fig:distillation}
\end{figure}

Model distillation includes two components: a teacher network and a student network. The teacher network is a well-trained established deep learning model (BEHRT in our case), and the student network is a simpler model or a target model that we intend to use it to distil knowledge from the teacher network (BDLD in our case). Tang et al.~\cite{tang2019distilling} developed a strategy to train a student model with partial assistance from teacher models to help the student network achieve a competitive performance, and Korattikara et al.~\cite{korattikara2015bayesian} proposed a Bayesian deep learning-based knowledge distillation method to transfer knowledge between probabilistic models. In this work, we combine both methods to use the probabilistic teacher network BEHRT to partially supervise the training of the probabilistic student network BDLD.  As shown in Figure~\ref{fig:distillation}, both the teacher network and the student network use the same input information. Afterwards, the prediction from the teacher network will be used as a soft label to partially supervise the training of the student model. At the same time, the student model is also trying to predict the real label. Therefore, the loss function can be summarized as below:
\begin{equation}
L=\alpha\cdot L_{ELBO}+(1-\alpha)\cdot L_{distill},
\end{equation}
where $\alpha$ represents a pre-set weight to balance the training objective for mimicking the teacher and leaning for the classification task. $L_{ELBO}$ represents the loss for evidence lower bound, and $L_{distill}$ is the loss for knowledge distillation. We used cross entropy for negative log-likelihood in evidence lower bound.  Additionally, the loss for knowledge distillation is shown as below:
\begin{equation}
L_{distill}= p(y|x,\boldsymbol{\theta^t})logS(y|x,\boldsymbol{\omega})+(1-p(y|x,\boldsymbol{\theta^t}))log(1-logS(y|x,\boldsymbol{\omega})),
\end{equation}
where $S(y|x,\boldsymbol{\omega})$ represents the outputs from the student network, $p(y|x,\boldsymbol{\theta^t})$ represents the outputs from the teacher network, $\boldsymbol{\theta^t}$ are the parameters for the teacher network, and $\boldsymbol{\omega}$ is the weights for student network. The distillation loss can also be considered as a reframing of the standard cross entropy loss.

\subsection{Population-wise risk association analyses}
The BDLD model decomposed the BEHRT model into two risk components: a contextual variable and an additive variable. The additive variable was the summation of coefficients of events that occurred in a patient’s medical history, therefore, the coefficients directly reflected the association between the corresponding event and predicted risk. 

\begin{equation}
Additive\; risk = \sum_{i=1}^{E} c_{i} \times e_{i},
\end{equation}

\noindent where c and e represent coefficient and event, respectively, and E represents events been considered by the multi-hot encoder. 

For the contextual variable, we assumed a well-trained BDLD model that learned the population-wise pattern and can be used as a risk estimator to evaluate risk over population. Therefore, we assessed the association of an event (e.g., a disease or a medication) to HF by calculating the relative average contextual ratio between a group of patients without the exposure and a group of patients with the exposure similar to estimating the risk ratio in epidemiological studies. We stratified the relative contextual ratio by age, and used the mean to represent the final relative contextual ratio. More details about age stratified calculation can be found in Appendix~\ref{sec:age_stratified}.

\begin{equation}
CR = \frac{P(outcome|exposure=False)}{P(outcome|exposure=True)},
\end{equation}
where CR represents relative contextual ratio, $P(outcome|exposure=True /False) = \frac{1}{N} \sum_{i=1}^{N}P(outcome|exposure=True/False,C_{i})$, $N$ represents the number of patients in the exposure group and non-exposure group accordingly, and $C_{i}$ represents the contextual information (records) for patient $i$. For each patient $i$, the risk $P$ is the average contextual variable over 30 samples.

We used this approach to identify events, more specifically diseases, that were positively and negatively associated with HF.

\subsection{Patient-wise feature selection}
\begin{figure}[h]
  \centering
  \includegraphics[width=\linewidth]{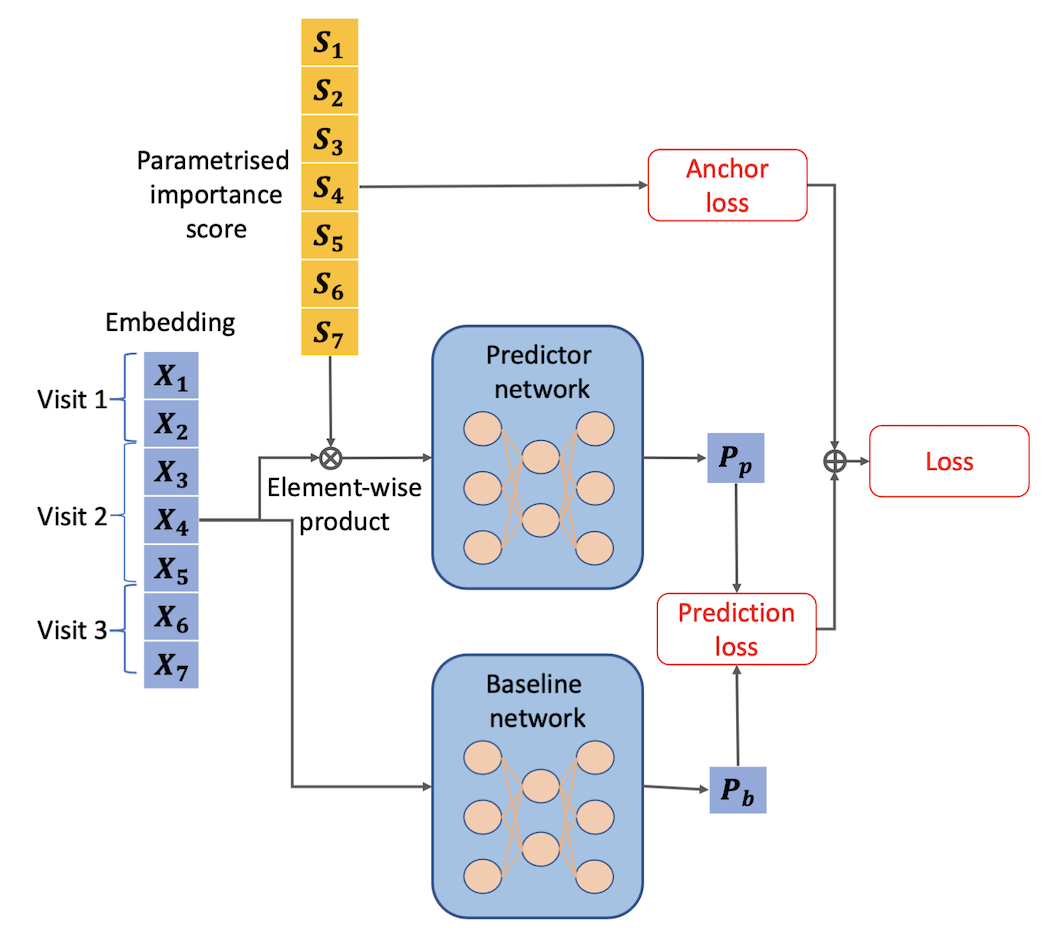}
  \caption{Block diagram of the patient-wise feature selection method. The figure shows the training mechanism for patient-wise feature selection. Embeddings, which are the summation of age and encounter embedding, are fed into both the baseline network and the predictor network. The baseline network thus provides a reference prediction $P_{b}$. Before feeding into the predictor network, the inputs are filtered with an importance score, therefore, it has a different prediction  $P_{p}$. The objective is to minimise the summation of prediction loss and anchor loss.}
  \label{fig:feature_selection}
\end{figure}

We proposed a novel patient-wise feature selection method to identify the most relevant event in a patient’s medical history for his/her prediction. It used a parametrized Gumbel-softmax distribution to learn an importance score (between 0 to 1) of each event. Our method consisted of 2 networks: a baseline network and a predictor network. The baseline model was a well-trained BDLD model and it predicts the risk using all medical records. We used such risk probability as reference to train the predictor model. For the predictor model, it was the same as the BDLD model with all the weights fixed but with an importance-score layer (Figure~\ref{fig:feature_selection}). Our optimization objective was to preserve the predictive probability from the predictor network as close as possible to the predictive probability from the baseline model and keep the number of records with high importance score as minimal as possible. Therefore, our proposed method was capable of discovering the most relevant features to support risk prediction. 

This approach took inspiration from  the concept of a local surrogate model~\cite{iml}, which optimizes an independent interpreter for individual prediction. In order to detect the personalized association for incident HF prediction, we implemented an importance score vector that assigned a trainable variable to each record. Each variable of the importance score vector is ideally a Bernoulli distribution, representing whether a record is important for the prediction or not. However, Bernoulli distribution is not differentiable. Therefore, we used the Gumbel-Softmax distribution~\cite{jang2017categorical} as an estimator for the Bernoulli variables (between 0 and 1), which is proved to be an efficient training strategy for Bernoulli and categorical variables~\cite{jang2017categorical}. The training strategy is to optimize the trainable importance score to preserve a model’s predictive capability while keeping as less important records as possible. We summarized this strategy into a loss function as below:

\begin{align} 
Loss= L_{prediction}+ \gamma \cdot L_{anchor},\\
L_{anchor}=SUM(\boldsymbol{\omega_{importance}}),\\
L_{prediction}=MSE(P_{b},P_{p}),
\end{align}

\noindent where $SUM(\boldsymbol{\omega_{importance}})$ represents summation over Bernoulli variables (importance score) across events, $MSE$ represents mean square error, and $\gamma$ represents weight coefficient (usually between 0 and 1).

\subsection{Experimental setup}
Models are implemented in PyTorch~\cite{NEURIPS2019_9015} and GPytorch~\cite{gardner2019gpytorch}. We followed the parameters been set for the probabilistic BEHRT model~\cite{li2020deep} with maximum sequence length 256, hidden size 150, 4 layers of transformer, 6 attention heads, 108 intermediate hidden size, and 24 pooling size. We used 40 inducing points and radial basis function (RBF) kernel for the Gaussian processes classifier. For BDL(D), we used similar hyper parameters. Both encounter and age embeddings as well as hidden size for BiLSTM were 150, and 256 was set for the maximum sequence length. For those components that are stochastic and deployed following the Bayesian deep learning framework, we used the mean-field distribution to approximate the posterior. The normal distribution with 0 mean and 0.374 standard deviation was used as prior distribution for the weights of those components. For the sparse Gaussian processes classifier in the Bayesian deep learning model, we used 100 inducing points and the RBF kernel. We used $\alpha=0.5$ for the model distillation training, and $\gamma=0.1$ for patient-wise feature selection. The Adam~\cite{kingma2017adam} optimizer was used for both BEHRT and BDL(D). For BDLD in model distillation, we used the learning rate $7e^{-4}$ and batch size 256. The number of training epochs was governed by an early stopping strategy,  and the training stopped when the loss did not improve after 5 epochs. For patient-wise feature selection,  we used the learning rate $9e^{-2}$, and each sample was trained for 500 iterations. 

\section{Results}
\subsection{Statistics for the risk prediction dataset}
The characteristics of the patients in the training set, tuning set, and validation set, as well as in all patients are summarized in Table~\ref{tab:patient}. All datasets were very similar with respect to demographics and statistics of the number of visits. 

\begin{table*}[h!]
  \caption{Characteristics of patients}
  \label{tab:patient}
  \begin{tabular}{ccccl}
    \toprule
     &All patients & Training set & Tuning set & Validation set\\
    \midrule
    General characteristics & & & & \\
    \midrule
    No. of patients (\% women) & 788,880 (63) & 473,394 (63) & 78,900 (63) & 236,586 (63) \\
    No. of diagnosis code & \multicolumn{4}{c}{1,533} \\
    No. of medication code & \multicolumn{4}{c}{360} \\
    \% White ethnicity (n) & 43.8 (345,515) & 43.8 (207,548) & 43.8 (34,567) & 43.7 (103,420)\\
    \% Unknown ethnicity (n) & 42.6 (336,092) & 42.6 (201,609) & 42.6 (33,613) & 42.6 (100,870)\\
    \midrule
    Baseline characteristics & & & & \\
    \midrule
    Mean No. of visits per patient & 84.9 & 85.0 & 84.9 & 84.8\\
    Mean No. of codes per visit & 2.02 & 2.02 & 2.02 & 2.02\\
    Mean No. of diagnosis codes per visit & 0.48 & 0.47 & 0.47 & 0.47 \\
    Mean (SD) follow-up (year) & 8.9 (5.2) & 8.9 (5.2) & 8.9 (5.2) & 8.9 (5.1)\\
    Mean (SD) age (year) & 57.1 (18.3) & 57.1 (18.2) & 57.1 (18.3)& 57.1 (18.3)\\
    \bottomrule
    SD: standard deviation\\
  \end{tabular}
\end{table*}

\subsection{Model performance evaluation}
We compared two models in three scenarios as described in Section~\ref{sec:evaluation}: A Transformer-based probabilistic model, BEHRT, a proposed model with simpler architecture trained with knowledge distillation mechanism (i.e., BDLD), and the same simple model but trained without knowledge distillation (i.e., BDL).  

\begin{figure}[h]
  \centering
  \includegraphics[width=\linewidth]{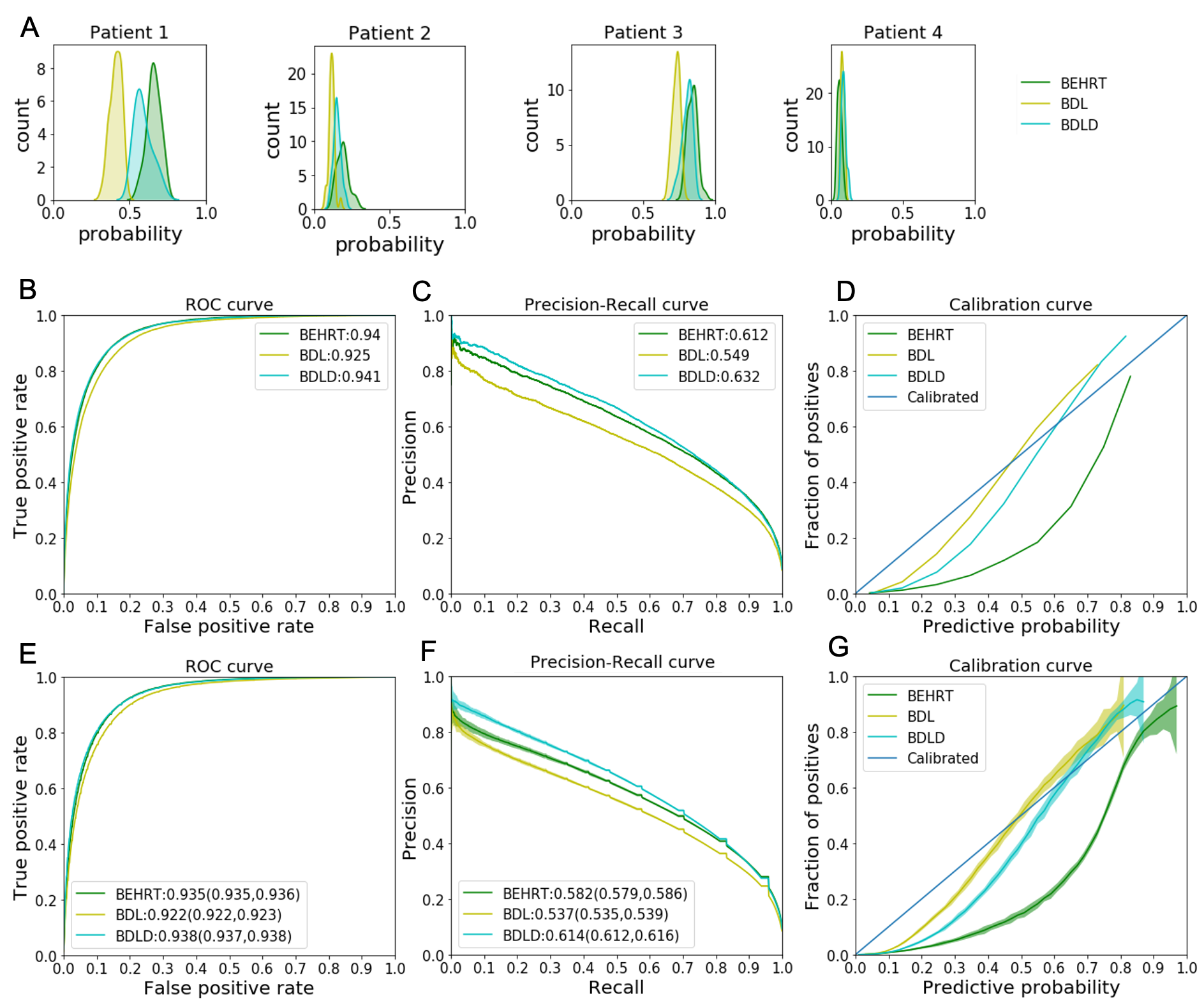}
  \caption{Performance evaluation of AI models on the validation set. BEHRT is an established incident HF risk prediction model, BDLD is our proposed Bayesian deep learning model trained with a distillation mechanism, and BDL is the same Bayesian deep learning model trained without distillation mechanism. A: Examples of predictions from four randomly selected patients, instead of predicting a single probability, all three models can present the predictive distribution, here, 30 samples are drawn from the predictive distribution for its approximation. B, C, D represent comparison of the ROC curve, precision-recall curve, and calibration curve among three models, all of them are conducted based on the mean predictive probability (i.e., average probability over 30 samples). E, F, G present the comparison of the ROC curve, precision-recall curve, and calibration curve by treating each sample separately, resulting in 30 sets of samples. Therefore, these figures can represent the uncertainty of the predictions with mean and 95\% confidence interval.}
  \label{fig:performance}
\end{figure}

The predictive performance of BEHRT, BDLD, and BDL on the validation set are shown in Figure~\ref{fig:performance}. Figure~\ref{fig:performance}A) exemplifies the probability distribution of developing HF, as predicted by the three models, for four randomly selected patients. In terms of average predictive probability (Figure~\ref{fig:performance}B, C, D), BDLD achieved the best performance, and substantially outperformed the BDL model, reaching 0.94 and 0.63 for AUROC and AUPRC, respectively. It showed a comparable performance to BEHRT with a slight improvement (2\%) in AUPRC. Additionally, BDLD also showed a better calibration than BEHRT. As similar pattern was observed when evaluating performance using mean with 95\% confidence interval over 30 samples from the predictive distribution (Figure~\ref{fig:performance}E, F, G). However, it is worth pointing out that the calibration curve in Figure~\ref{fig:performance}G shows that BDLD preserved a similar uncertainty estimation pattern as BEHRT. Both of them indicated that the risk predictions at the higher end were more uncertain in our highly imbalanced dataset (with only 8.3\% of positive samples).

In general, the model performance comparison illustrated that the proposed BDLD model achieved a similar performance to BEHRT despite its much simpler architecture. Moreover, BDLD preserved a similar uncertainty pattern as the BEHRT model, therefore, supporting its potential to replace the original BEHRT model for more down-stream risk association analysis.

\subsection{Relation between latent variables and risk for BDLD}
\begin{figure}[h]
  \centering
  \includegraphics[width=\linewidth]{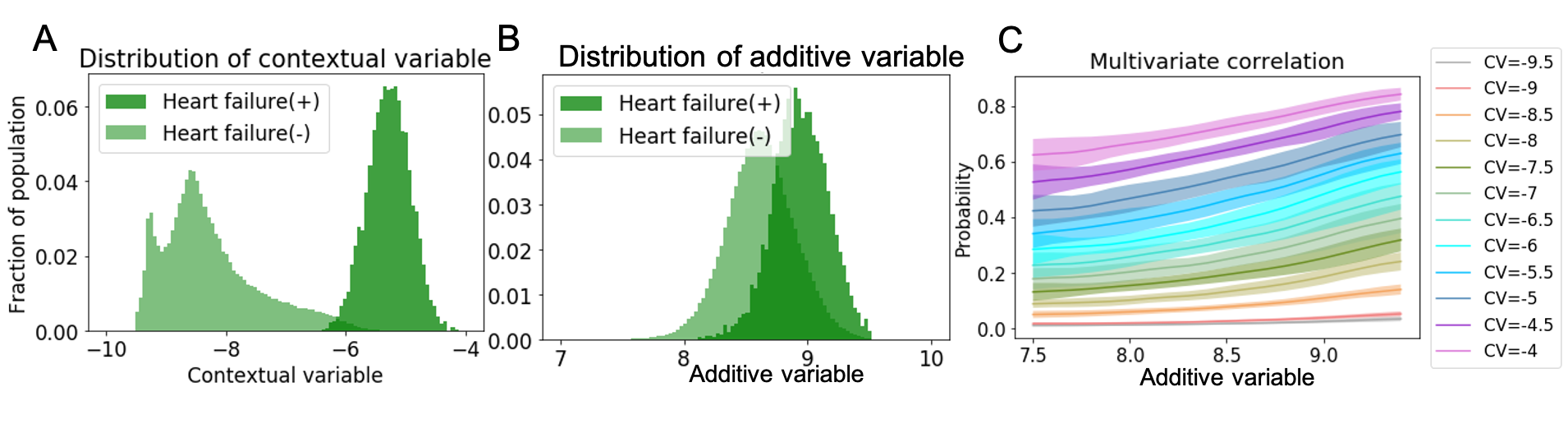}
  \caption{Analyses for multivariate correlation among contextual variable, additive variable, and predicted risk. A and B represent distribution of the contextual and additive variable, respectively. C is the multivariate correlation with the x axis represents additive variable, color represent contextual variable, and y axis represents the predictive distribution of HF risk}
  \label{fig:multivariate}
\end{figure}

Our proposed BDLD model used a GP classifier to extract the multivariate correlation among contextual variables, additive variables, and predicted (HF) risk (Figure~\ref{fig:model_arch}). The contextual variable captured the temporal interactions of input features using a BiLSTM, and the additive variable was the weighted sum of input features representing diagnoses and medication that ever occurred in a patient’s history. Both variables together indicated a patient’s risk of developing HF. Figure~\ref{fig:multivariate}A and \ref{fig:multivariate}B show that the range of the contextual variable for HF (+) and HF (-) are approximately between -9.5 and -4, and the approximate range for the additive variable is between 7.5 to 9.5. We also see that the contextual variable is better at distinguishing between positive and negative cases while there remains substantial overlap when additive variables are considered, and combining both risk variables can lead to more reliable risk prediction (Figure~\ref{fig:multivariate}C). We also need to clarify here that both variables are latent features learned by a neural network, and they are not numerically meaningful. We are more interested in understanding how these two variable groups relate to the risk of HF. 

By considering the range of contextual variables and additive variables simultaneously, Figure~\ref{fig:multivariate}C further demonstrates the multivariate correlation between those two variables and the predicted risk. It shows that (1) the higher the value of the contextual variable and the higher the value of the additive variable, the higher the predicted risk, (2) high risk predictions have higher uncertainty than lower risk predictions, and (3) when the contextual variable is low (more towards -9.5), the prediction is extremely certain, and the additive variable does not provide much added value. We build on these trends as the basis of the population-wise association analysis in the next section.

\subsection{Population-wise association analysis}\label{sec:pop_association}
\begin{figure}[h]
  \centering
  \includegraphics[width=\linewidth]{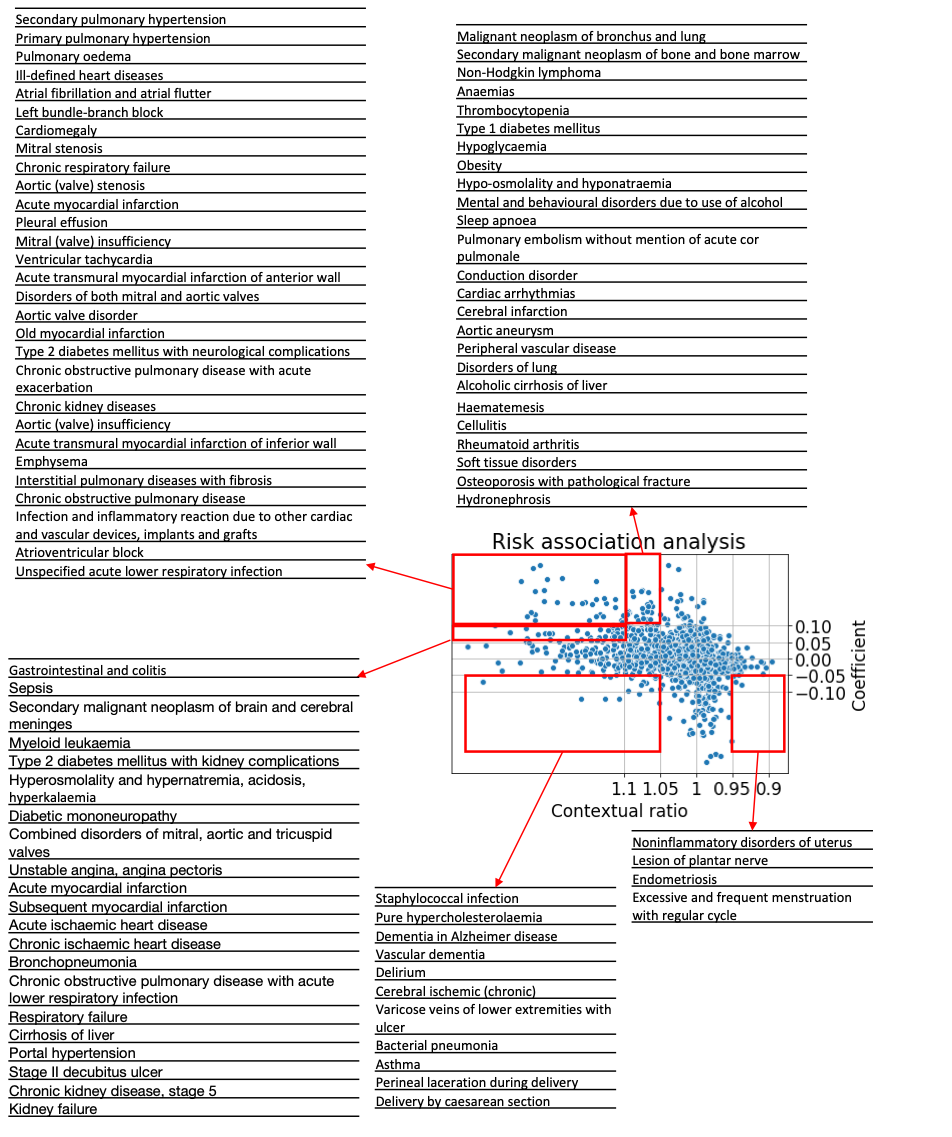}
  \caption{Risk factor association analysis. The x axis represents the contextual ratio and the y axis represent coefficient of diseases; the higher the contextual ratio and the higher the coefficient, the stronger the association between a disease and HF. The red boxes are randomly selected for visualization.}
  \label{fig:association}
\end{figure}

By calculating the contextual ratio of the non-exposure group over the exposure group for diseases, along with their coefficients (additive variable is summation of coefficient of the diseases that occurred in a patient’s medical records), we conducted analyses to identify population-wise risk associations (Figure~\ref{fig:association}). As previous analyses demonstrated, the higher the contextual variable and additive variable, the higher the predicted risk. Therefore, events (i.e., diseases) that have a higher contextual ratio and higher coefficient will have stronger associations with HF. Following this criterion, we identified 598 diseases with contextual ratio higher than 1 and coefficient higher than 0, indicating that they are potential associations. As shown in Figure~\ref{fig:association}, the upper left corner shows diseases with strong associations with HF, and most of them appear to be established risk factors, supporting the effectiveness of our proposed methods. Furthermore, we also identified 379 diseases that had a contextual ratio lower than 1 and a coefficient lower than 0, indicating potential dissociations. Although the signal for dissociations was weak (diseases do not have a low contextual ratio and low coefficient at the same time),  comparing to the signal for associations, it is still worth pointing out that a few conditions such as ‘noninflammatory disorders of uterus’, ‘lesion of plantar nerve’, ‘endometriosis’, and ‘excessive and frequent menstruation with regular cycle’ indeed showed relatively strong dissociations with HF (suggesting that these conditions are associated with lower risk of HF). We also included an age distribution analysis in Appendix~\ref{sup:age_analysis} for the identified dissociations to ensure they were conducted in a population with a reasonable age distribution rather than analyzing population with extreme cases. However, of course, these hypotheses require further confirmation in other studies or clinical trials. For diseases with contextual ratio higher than 1 and coefficient lower than 0 (lower left quadrant), the interpretation is unclear. One potential interpretation is that patients with those diseases indeed have a higher risk of developing HF, however, they have a far less direct association with HF, thus, increasing the uncertainty for the predicted risk. A similar assumption can also be applied to diseases with contextual ratio lower than 1 and coefficient higher than 0 (upper right quadrant). We also encourage readers to read the discussion in Appendix~\ref{sup:correlation} regarding potential concerns about correlation (or collinearity) among independent predictors. 

\subsection{Patient-wise association analysis}
\begin{figure}[h]
  \centering
  \includegraphics[width=\linewidth]{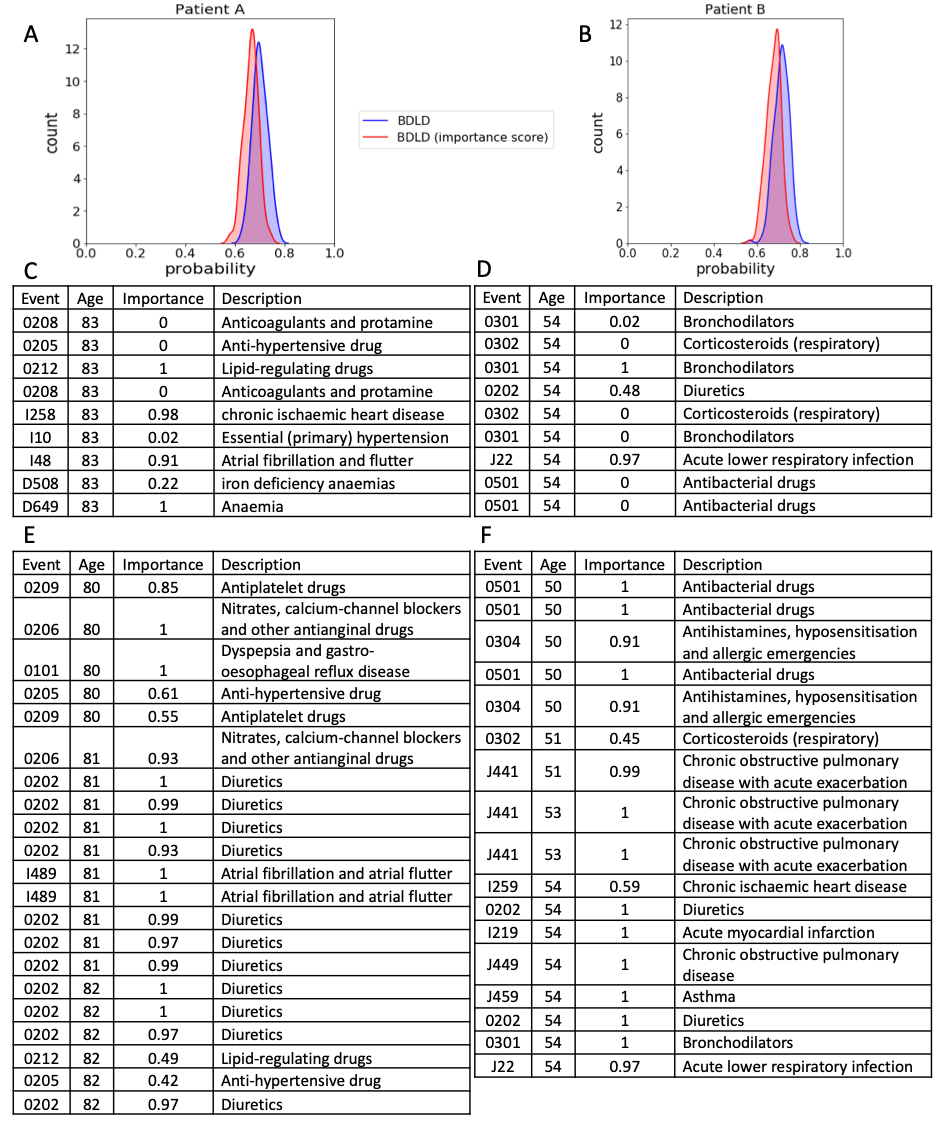}
  \caption{Patient-wise feature selection to identify associations and provide interpretable explanations of the predictions. Two patients are used for the illustration, with only a fraction of records been shown here. A and B represent the model predictive distributions for the model using full medical records, and a fraction of records with high importance score for patients A and B, respectively.   C and D show a fraction of records for patients A and B. Event is event code, age is the age when corresponding event occurs, and the importance score is a trainable score between 0 and 1 to indicate the importance of the event, where 1 indicates high importance and 0 indicates low importance. E and F list events with relatively high importance score for patients A and B respectively.}
  \label{fig:patient_analysis}
\end{figure}

In many cases, especially when patients are highly heterogeneous, clinicians can be more interested in identifying patient-wise associations to support decision-making. Our proposed variable selection method to identify the importance score of each longitudinal record can untangle complex medical records and provide an interpretable explanation of the predictions made. As shown in Figure~\ref{fig:patient_analysis}, we showcased two random examples and only a fraction of records is listed for illustration (full records can be found in Appendix~\ref{sup:reords}). Figure~\ref{fig:patient_analysis}A and \ref{fig:patient_analysis}B firstly show that for both patients, only using a small fraction of key events (with high importance score), the model (BDLD) can achieve a similar predictive performance as using a patient’s entire medical history. It supports the effectiveness of the identified key events. Furthermore, Figures~\ref{fig:patient_analysis}C and \ref{fig:patient_analysis}D show a small fraction of records for both patients A and B. They demonstrate that the proposed method is very dynamic and can adapt to different patient scenarios to identify personalized key events depending on their context. We summarized the events with relatively high importance score for both patients A and B in Figures~\ref{fig:patient_analysis}E and ~\ref{fig:patient_analysis}F, respectively. For patient A, we see that ‘antiplatelet drugs’, ‘nitrates, calcium-channel blockers and other antianginal drugs’, ‘dyspepsia and gastro-oesophageal reflux disease’, ‘diuretics’, and ‘atrial fibrillation and atrial flutter’ substantially contribute to the HF prediction. Because of the intensive repetition of ‘diuretics’ and most of its repetition are identified with a high importance score, we know ‘diuretics’ plays an important role for patient A. For patient B, a quite different pattern is identified. We see that ‘antibacterial drugs’, ‘antihistamines, hypersensitisation and allergic emergencies’, ‘chronic obstructive pulmonary disease with acute exacerbation’, ‘diuretics’, ‘acute myocardial infarction’, ‘asthma’, ‘bronchodilators’, and ‘acute lower respiratory infection’ are important key events that lead to the high risk prediction.  

\subsection{Discussion}
Although deep learning models have been widely applied to various prognosis tasks and showed a superior predictive performance than the conventional statistical models, there still are major concerns about their explainability, which is becoming a major bottleneck that hinders the wider application of deep learning models for clinical usage and research. To tackle this difficulty, in this study, we provided a proof of concept that model distillation can be used as a discovery-enabling tool to identify population-wise potential risk factors for HF from a data-driven way without explicit expert engagement, and a variable selection analysis can accurately identify patient-wise associations to provide interpretable explanations to support trustworthiness of model predictions and decision-making. Although our work mainly focused on HF, the methodologies carried out from this work can be easily applied to other diseases of interest with minimal changes. 

In this work, we introduced several methodological innovations for deep learning model interpretation. Firstly, instead of directly interpreting a deep and complex model (i.e., BEHRT), we used a model distillation mechanism to decompose the pattern learned by the deep model into two simpler components. One component represented the hidden interactions within the temporal trajectory, and the other component represented the risk of each disease or medication to HF. We carefully designed these two components to be alike to the concept of interaction and independent predictor in the conventional statistical model for better interpretation. Despite the simple architecture of the distilled model, it was capable of achieving comparable or even better predictive performance than the BEHRT model, while preserving the uncertainty information which suggested that HF (+) predictions were more uncertain than HF (-) predictions in our highly imbalanced dataset. 

Secondly, to untangle the risk of a disease in its temporal context, we combined the concept of relative risk and partial dependent plot~\cite{iml}, using ithe nteraction component as an estimator and marginalizing the risk of patients over the exposure group and the non-exposure group to evaluate the relative risk of a disease. By using such a relative risk along with the learned coefficient of a disease, we managed to conduct an association map and saw that our findings were broadly consistent with prior knowledge of HF and corroborated risk factors of the disease. Our association map also raises questions that warrants further research. For instance, although there is some evidence from traditional epidemiological studies for an association between asthma and HF~\cite{sun2017history}, this could not be supported in our study. Furthermore, as dissociation is a less well-established area in risk analysis, we surprisingly found potential dissociation between menstruation with regular cycle and HF aligned with the hypothesis of an on-going trial~\cite{menstrual}. Therefore, our proposed method might be of great use to provide potential candidates for the investigation of factors that prevent a particular disease.

Thirdly, although population-wise risk association plays an important role in disease prevention, in practice, clinicians can be more interested in identifying critical events for individual patients to support decision-making. Our proposed variable selection method is designed to tackle this demand, showing a promising predictive performance even with only a small fraction of critical records. More specifically, we saw that using a subset of records with high importance score for prediction, the predictive distribution significantly aligned with the predictive distribution with all records, which proved the effectiveness of those key events. Additionally, our proposed variable selection method is very dynamic and can adapt to patients who are heterogeneous. For instance, in our results, we listed two patients who have different age and different trajectory (medical history). Our model indicated that ‘diuretics’ was an important signal for the older patient. Since diuretics are usually prescribed to patients who have symptoms of HF, it is reasonable that the model identifies it as a predictor. However, for the other younger patient, more diverse signals are captured. It shows a combination of chronic obstructive pulmonary disease with acute exacerbation, diuretics, acute myocardial infarction, asthma, bronchodilators, and acute lower respiratory infection in this patient’s recent history renders the patient high risk. By simplifying the medical records and emphasising the most important events, this proposed tool can greatly untangle the complexity of EHR and provide evidence to support prediction and aid clinical decision making, in particular when disease phenotypes are heterogeneous.

In general, we believe our proposed methods have a number of implications. First, model distillation can significantly prune the model and reduce model complexity, yet preserve the deep, complex model’s predictive capability. Thus, the deep learning model can be easily deployed to various scenarios (e.g., limited computational and storage capacity, low latency) to aid decision-making process. For instance, the small model can be deployed on a local clinical environment or smart phone where computational resources are limited. Because of the accessibility of the model, it can be used to alert clinicians and even patients themselves for a presentative check when their longitudinal medical records are available. Secondly, our distillation model can distil the knowledge learned by a complex deep learning model into a simple model. By interpreting the contextual and independent pattern, it can provide a data-driven perspective to rapidly identify risk factors in both population and individual level without explicit engagement from experts.

Our work also has limitations. First, we only used limited modalities (diagnoses, medications, and age) for modelling. By incorporating more available information from EHR, we can potentially improve the predictive performance for HF prediction, leading to more accurate measurement for contextualised dependencies and providing a calibrated measurement for independent associations. Moreover, we only tested our methods on one EHR; all conclusions highly depend on our dataset. Therefore, further research must be pursued regarding transferability to other datasets.


\bibliographystyle{ACM-Reference-Format}
\bibliography{sample-base}

\newpage
\appendix
\section{Research Methods}

\subsection{Age stratified relative contextual risk analyses}\label{sec:age_stratified}
To ensure the fairness for the calculation of risk ratio, we carefully conducted an age-stratified subgroup selection. For each disease, we grouped patients based on their age of the first incidence and their baseline age. The age groups that we considered were 45-55, 55-60, 60-65, 65-70, 70-75, 75-80, 80-85, and 85-90. For example, if diabetes is our interest, we will choose patients who have the incidence of diabetes between 45-55 (e.g., 46) and baseline between 45-55 (e.g., 54) in the exposure group, and choose patients who have never been diagnosed with diabetes before baseline age (between 45-55, e.g. 54) in the non-exposure group. We repeated this estimation for 45-55/55-60 (incident age/baseline age), 45-55/60-65, 45-55/65-70, 45-55/70-75 and other possible combinations, we ignored any group pair if there are no patients in either exposure group or non-exposure group. In the end, we used the mean over all age groups as the summarized single value to represent a disease’s relative contextual ratio.

\begin{figure}[h]
  \centering
  \includegraphics[width=\linewidth]{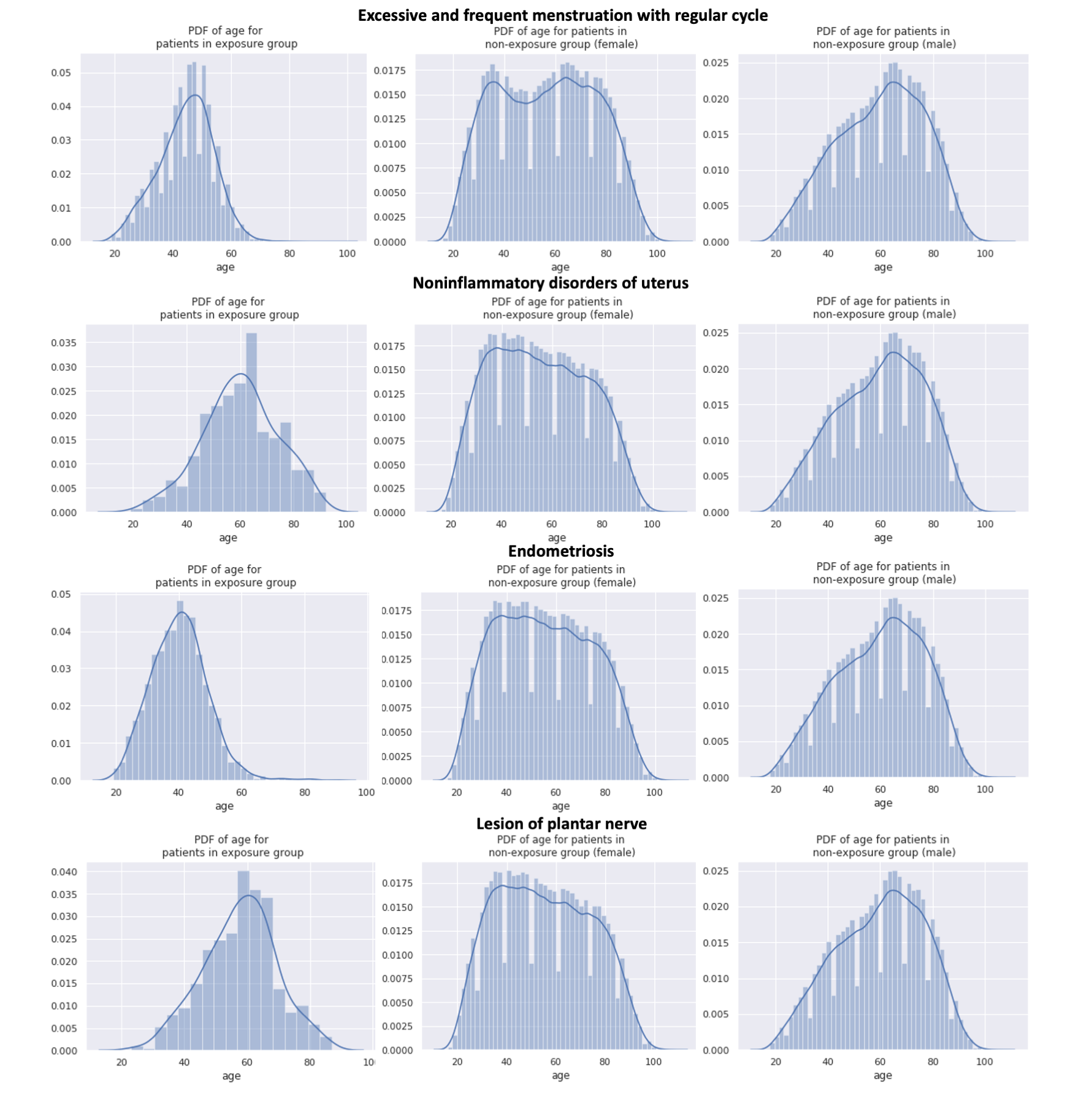}
  \caption{Baseline age distribution of exposure group and non-exposure group for diseases with disassociation. The left column represents age distribution for patients in the exposure group, the middle column represents age distribution of female patients in the non-exposure group, and the right column represents age distribution of male patients in the non-exposure group. PDF represents probability density function.}
  \label{fig:age_analysis}
\end{figure}

\section{Supplementary results}
\subsection{Age distribution analysis for diseases with disassociation}\label{sup:age_analysis}

As shown in Figure~\ref{fig:age_analysis}, we see for all four diseases that are identified as disassociations, the distribution of baseline age in exposure group in general covers a quite wide range of ages with median between 40 to 60. Since we stratified our association analysis by age and the exposure groups are not extreme cases, the conclusion from our association analysis should be valid.

\subsection{Discussion about correlation among independent predictors}\label{sup:correlation}
The distilled-model used a one-layer feed-forward network (linear component) to learn the coefficient of each event. To address the concern for feature correlation, we randomly sampled 100,000 patients in the training set to calculate the Cramér’s correlation between each feature pair. Among 1,898 events, only 19 pairs have correlation higher than 0.6 (Table~\ref{tab:correlation}). Additionally, considering we used variational inference with standard Gaussian prior for the weight training, which equals to adding L2-normal regulariser to penalise the training process, we would not worry about the feature correlation problem. 

\begin{table}[h!]
  \caption{Cramér’s correlation for disease/medication pair with correlation higher than 0.6}
  \label{tab:correlation}
  \begin{tabular}{ccl}
    \toprule
     Event A&	Event B&	Correlation\\
    \midrule
    D57.0&	D57.1&	0.70 \\
    B37.3&	N77.1&	0.94 \\
    L40.5&	M07.3&	0.92 \\
    M51.1&	G55.1&	0.82 \\
    C15.9&	C15.5&	0.60 \\
    0106/0704/1002&	G70.0&	0.75\\
    0602/6407&	E05.9&	0.68\\
    7408&	7411&	0.60\\
    7109&	7408&	0.60\\
    7409&	7408&	0.79\\
    1306&	L70.0&	0.62\\
    1305&	L40.9&	0.62\\
    1103&	H10.3&	0.63\\
    7154&	7119&	0.75\\
    1106&	H40.9&	0.60\\
    0602&	E03.9&	0.74\\
    0601&	E11.9&	0.63\\
    0601&	7119&	0.64\\
    0601&	7154&	0.71\\
    \bottomrule 
  \end{tabular}
\end{table}

\subsection{Patient-wise association identification}\label{sup:reords}
Here, we list the full records for two random selected patients (i.e., Patient A and B) with the importance scores. Patients A and B are shown in Tables~\ref{tab:patienta} and~\ref{tab:patientb}, respectively.

\captionof{table}{Patient A}
\label{tab:patienta}
\tablefirsthead{\toprule Code&Age&Score&Description \\ \midrule}
\begin{supertabular}{|p{0.22\linewidth}|p{0.05\linewidth}|p{0.07\linewidth}|p{0.5\linewidth}|}
 0212&80&0&Lipid-regulating drugs\\
0212&80&0&Lipid-regulating drugs\\
0209&80&0.85&Antiplatelet drugs\\
0206&80&1&Nitrates, calcium-channel blockers and other antianginal drugs\\
0101&80&1&Dyspepsia and gastro-oesophageal reflux disease\\
0205&80&0.61&Anti-hypertensive drug\\
0209&80&0.55&Antiplatelet drugs\\
0212&80&0.05&Lipid-regulating drugs\\
0205&80&0&Anti-hypertensive drug\\
0101&80&0.01&Dyspepsia and gastro-oesophageal reflux disease\\
0206&80&0&Nitrates, calcium-channel blockers and other antianginal drugs\\
0101&81&0&Dyspepsia and gastro-oesophageal reflux disease\\
0212&81&0&Lipid-regulating drugs\\
0209&81&0&Antiplatelet drugs\\
0206&81&0&Nitrates, calcium-channel blockers and other antianginal drugs\\
0205&81&0.05&Anti-hypertensive drug\\
J220&81&0.01&Acute lower respiratory infection\\
0501&81&0&Antibacterial drugs\\
0209&81&0.08&Antiplatelet drugs\\
0205&81&0&Anti-hypertensive drug\\
0101&81&0&Dyspepsia and gastro-oesophageal reflux disease\\
0206&81&0&Nitrates, calcium-channel blockers and other antianginal drugs\\
0212&81&0&Lipid-regulating drugs\\
J220&81&0&Acute lower respiratory infection\\
0501/0504/1306&81&0&Antibacterial drugs/Antiprotozoal drugs/Acne and rosacea\\
0212&81&0&Lipid-regulating drugs\\
0209&81&0&Antiplatelet drugs\\
0206&81&0&Nitrates, calcium-channel blockers and other antianginal drugs\\
0101&81&0&Dyspepsia and gastro-oesophageal reflux disease\\
0205&81&0&Anti-hypertensive drug\\
0501/0504/1306&81&0&Antibacterial drugs/Antiprotozoal drugs/Acne and rosacea\\
0205&81&0.01&Anti-hypertensive drug\\
0212&81&0&Lipid-regulating drugs\\
0101&81&0&Dyspepsia and gastro-oesophageal reflux disease\\
0209&81&0&Antiplatelet drugs\\
0206&81&0&Nitrates, calcium-channel blockers and other antianginal drugs\\
0101&81&0&Dyspepsia and gastro-oesophageal reflux disease\\
0205&81&0&Anti-hypertensive drug\\
0209&81&0&Antiplatelet drugs\\
0212&81&0&Lipid-regulating drugs\\
0206&81&0.93&Nitrates, calcium-channel blockers and other antianginal drugs\\
0202&81&1&Diuretics\\
0206&81&0&Nitrates, calcium-channel blockers and other antianginal drugs\\
0205&81&0.37&Anti-hypertensive drug\\
0101&81&0&Dyspepsia and gastro-oesophageal reflux disease\\
0209&81&0&Antiplatelet drugs\\
0202&81&1&Diuretics\\
0212&81&0&Lipid-regulating drugs\\
0101&81&0&Dyspepsia and gastro-oesophageal reflux disease\\
0205&81&0&Anti-hypertensive drug\\
0206&81&0&Nitrates, calcium-channel blockers and other antianginal drugs\\
0209&81&0&Antiplatelet drugs\\
0212&81&0.01&Lipid-regulating drugs\\
0202&81&0.01&Diuretics\\
0209&81&0&Antiplatelet drugs\\
0206&81&0&Nitrates, calcium-channel blockers and other antianginal drugs\\
0212&81&0&Lipid-regulating drugs\\
0202&81&0.99&Diuretics\\
0101&81&0&Dyspepsia and gastro-oesophageal reflux disease\\
0205&81&0&Anti-hypertensive drug\\
0106&81&0&Laxatives\\
0101&81&0&Dyspepsia and gastro-oesophageal reflux disease\\
0212&81&0&Lipid-regulating drugs\\
0206&81&0&Nitrates, calcium-channel blockers and other antianginal drugs\\
0205&81&0&Anti-hypertensive drug\\
0202&81&1&Diuretics\\
0209&81&0&Antiplatelet drugs\\
0212&81&0&Lipid-regulating drugs\\
0205&81&0&Anti-hypertensive drug\\
0202&81&0.93&Diuretics\\
I489&81&1&Atrial fibrillation and atrial flutter\\
I489&81&1&Atrial fibrillation and atrial flutter\\
0202&81&0.99&Diuretics\\
0101&81&0&Dyspepsia and gastro-oesophageal reflux disease\\
0212&81&0&Lipid-regulating drugs\\
0208&81&0&Anticoagulants and protamine\\
0205&81&0&Anti-hypertensive drug\\
0212&81&0&Lipid-regulating drugs\\
0205&81&0&Anti-hypertensive drug\\
0208&81&0&Anticoagulants and protamine\\
0202&81&0.97&Diuretics\\
0206&81&0&Nitrates, calcium-channel blockers and other antianginal drugs\\
0208&81&0&Anticoagulants and protamine\\
0101&81&0&Dyspepsia and gastro-oesophageal reflux disease\\
0205&81&0&Anti-hypertensive drug\\
0212&81&0&Lipid-regulating drugs\\
0205&81&0&Anti-hypertensive drug\\
0212&81&0&Lipid-regulating drugs\\
0208&81&0&Anticoagulants and protamine\\
0202&81&0.99&Diuretics\\
0101&81&0&Dyspepsia and gastro-oesophageal reflux disease\\
0202&82&0.98&Diuretics\\
0212&82&0&Lipid-regulating drugs\\
0101&82&0&Dyspepsia and gastro-oesophageal reflux disease\\
0205&82&0&Anti-hypertensive drug\\
0208&82&0&Anticoagulants and protamine\\
0212&82&0&Lipid-regulating drugs\\
0101&82&0&Dyspepsia and gastro-oesophageal reflux disease\\
0208&82&0&Anticoagulants and protamine\\
0205&82&0&Anti-hypertensive drug\\
0202&82&1&Diuretics\\
0208&82&0&Anticoagulants and protamine\\
0212&82&0&Lipid-regulating drugs\\
0205&82&0&Anti-hypertensive drug\\
0202&82&1&Diuretics\\
0206&82&0&Nitrates, calcium-channel blockers and other antianginal drugs\\
0101&82&0&Dyspepsia and gastro-oesophageal reflux disease\\
I340&82&0.04&Mitral (valve) insufficiency\\
0208&82&0&Anticoagulants and protamine\\
0205&82&0&Anti-hypertensive drug\\
0212&82&0&Lipid-regulating drugs\\
0206&82&0&Nitrates, calcium-channel blockers and other antianginal drugs\\
0205&82&0&Anti-hypertensive drug\\
0208&82&0&Anticoagulants and protamine\\
0212&82&0&Lipid-regulating drugs\\
0202&82&0.97&Diuretics\\
0212&82&0.49&Lipid-regulating drugs\\
0202&82&0&Diuretics\\
0205&82&0&Anti-hypertensive drug\\
0101&82&0&Dyspepsia and gastro-oesophageal reflux disease\\
0208&82&0&Anticoagulants and protamine\\
K590&82&0&Constipation\\
0106&82&0&Laxatives\\
0212&82&0&Lipid-regulating drugs\\
0202&82&1&Diuretics\\
0208&82&0&Anticoagulants and protamine\\
0208&82&0&Anticoagulants and protamine\\
0212&82&0&Lipid-regulating drugs\\
0205&82&0.42&Anti-hypertensive drug\\
0202&82&0.97&Diuretics\\
0208&82&0&Anticoagulants and protamine\\
0212&82&0&Lipid-regulating drugs\\
0205&82&0&Anti-hypertensive drug\\
0208&82&0&Anticoagulants and protamine\\
0212&82&0&Lipid-regulating drugs\\
0202&82&0.31&Diuretics\\
0205&82&0&Anti-hypertensive drug\\
0208&82&0&Anticoagulants and protamine\\
0212&82&0&Lipid-regulating drugs\\
0202&82&0&Diuretics\\
0205&82&0&Anti-hypertensive drug\\
0202&82&1&Diuretics\\
0212&82&0&Lipid-regulating drugs\\
0208&82&0&Anticoagulants and protamine\\
0205&82&0&Anti-hypertensive drug\\
0205&83&1&Anti-hypertensive drug\\
0212&83&0&Lipid-regulating drugs\\
0208&83&0&Anticoagulants and protamine\\
0106&83&0&Laxatives\\
K590&83&0&Constipation\\
0205&83&0&Anti-hypertensive drug\\
0206&83&0.01&Nitrates, calcium-channel blockers and other antianginal drugs\\
0202&83&1&Diuretics\\
0212&83&0.02&Lipid-regulating drugs\\
0208&83&0&Anticoagulants and protamine\\
0101&83&0&Dyspepsia and gastro-oesophageal reflux disease\\
0101&83&0&Dyspepsia and gastro-oesophageal reflux disease\\
0202&83&1&Diuretics\\
0206&83&0&Nitrates, calcium-channel blockers and other antianginal drugs\\
0208&83&0.02&Anticoagulants and protamine\\
0212&83&0&Lipid-regulating drugs\\
0205&83&0&Anti-hypertensive drug\\
K590&83&0.14&Constipation\\
0106&83&0&Laxatives\\
0208&83&0&Anticoagulants and protamine\\
0205&83&0.01&Anti-hypertensive drug\\
0206&83&0.01&Nitrates, calcium-channel blockers and other antianginal drugs\\
0202&83&0.98&Diuretics\\
0212&83&0.03&Lipid-regulating drugs\\
0205&83&1&Anti-hypertensive drug\\
0208&83&0&Anticoagulants and protamine\\
0212&83&0.92&Lipid-regulating drugs\\
0205&83&0&Anti-hypertensive drug\\
0208&83&0&Anticoagulants and protamine\\
0212&83&0&Lipid-regulating drugs\\
0208&83&0&Anticoagulants and protamine\\
0212&83&0.03&Lipid-regulating drugs\\
0205&83&0&Anti-hypertensive drug\\
0202&83&0.09&Diuretics\\
0205&83&0&Anti-hypertensive drug\\
0212&83&1&Lipid-regulating drugs\\
0208&83&0&Anticoagulants and protamine\\
0205&83&0&Anti-hypertensive drug\\
0212&83&1&Lipid-regulating drugs\\
0208&83&0&Anticoagulants and protamine\\
I258&83&0.98&chronic ischaemic heart disease\\
I10&83&0.02&Essential (primary) hypertension\\
I48&83&0.91&Atrial fibrillation and flutter\\
D508&83&0.22&iron deficiency anaemias\\
D649&83&1&Anaemia\\
W19&83&0.31&Accidental fall\\
1003&83&0&Drugs for soft-tissue disorders and topical pain relief\\
1003&83&0&Drugs for soft-tissue disorders and topical pain relief\\
0212&83&0&Lipid-regulating drugs\\
0205&83&0&Anti-hypertensive drug\\
0103&83&0&Antisecretory drugs and mucosal protectants\\
0407&83&0&Analgesics\\
I259&83&1&Chronic ischaemic heart disease\\
I48&83&0.07&Atrial fibrillation and flutter\\
W199&83&1&Fall home during unspecified activity\\
W060&83&0.06&Fall from bed\\
 \bottomrule
\end{supertabular}

\captionof{table}{Patient B}
\label{tab:patientb}
\tablefirsthead{\toprule Code&Age&Score&Description \\ \midrule}
\begin{supertabular}{|p{0.22\linewidth}|p{0.05\linewidth}|p{0.07\linewidth}|p{0.5\linewidth}|}
J441&50&0.1&Chronic obstructive pulmonary disease with acute exacerbation\\
0304&50&0&Antihistamines, hyposensitisation and allergic emergencies\\
J301&50&0.04&Allergic rhinitis due to pollen\\
0105/0302/ 0501/0504/ 0603/0802/ 1001&50&0.01&Chronic bowel disorders/Corticosteroids (respiratory)/ Antibacterial drugs/Antiprotozoal drugs/ Corticosteroids (endocrine)/Drugs affecting the immune response/ Drugs used in rheumatic diseases and gout\\
0501&50&1&Antibacterial drugs\\
0301&50&0&Bronchodilators\\
F101&50&0.05&Mental and behavioural disorders due to psychoactive substance use (hamful use)\\
0304&50&0.03&Antihistamines, hyposensitisation and allergic emergencies\\
F102&50&0&Mental and behavioural disorders due to psychoactive substance use (dependence syndrome)\\
0501&50&1&Antibacterial drugs\\
0301&50&0&Bronchodilators\\
0301&50&0&Bronchodilators\\
0304&50&0.91&Antihistamines, hyposensitisation and allergic emergencies\\
0105/0302/ 0501/0504 /0603/0802/ 1001&50&0&Chronic bowel disorders/Corticosteroids (respiratory)/ Antibacterial drugs/Antiprotozoal drugs/ Corticosteroids (endocrine)/Drugs affecting the immune response/ Drugs used in rheumatic diseases and gout\\
J441&50&0.01&Chronic obstructive pulmonary disease with acute exacerbation\\
0501&50&0&Antibacterial drugs\\
0501&50&0&Antibacterial drugs\\
0301&50&0&Bronchodilators\\
0301&50&0&Bronchodilators\\
0501&50&0&Antibacterial drugs\\
J40&50&0&Bronchitis\\
0301&50&0&Bronchodilators\\
0301&50&0&Bronchodilators\\
0501&50&0&Antibacterial drugs\\
J029&50&0&Acute bronchitis\\
0301&50&0&Bronchodilators\\
0501&51&0&Antibacterial drugs\\
J40&51&0&Bronchitis\\
0501&51&0&Antibacterial drugs\\
0301&51&0&Bronchodilators\\
J42&51&0&chronic bronchitis\\
0302&51&0&Corticosteroids (respiratory)\\
0301&51&0&Bronchodilators\\
302&51&0&Corticosteroids (respiratory)\\
0302&51&0&Corticosteroids (respiratory)\\
0301&51&0&Bronchodilators\\
J40&51&0&Bronchitis\\
0501&51&0&Antibacterial drugs\\
0302&51&0&Corticosteroids (respiratory)\\
0301&51&0.01&Bronchodilators\\
0302&51&0&Corticosteroids (respiratory)\\
0301&51&0&Bronchodilators\\
0301&51&0&Bronchodilators\\
0304&51&0&Antihistamines, hyposensitisation and allergic emergencies\\
0302&51&0.45&Corticosteroids (respiratory)\\
J441&51&0.99&Chronic obstructive pulmonary disease with acute exacerbation\\
0501&51&0.05&Antibacterial drugs\\
0304&51&0&Antihistamines, hyposensitisation and allergic emergencies\\
0301&51&0&Bronchodilators\\
0105/0302 /0501/0504 /0603/0802 /1001&51&0&Chronic bowel disorders/Corticosteroids (respiratory)/ Antibacterial drugs/Antiprotozoal drugs/ Corticosteroids (endocrine)/Drugs affecting the immune response/ Drugs used in rheumatic diseases and gout\\
0302&51&0&Corticosteroids (respiratory)\\
0302&51&0&Corticosteroids (respiratory)\\
0304&51&0&Antihistamines, hyposensitisation and allergic emergencies\\
0301&51&0&Bronchodilators\\
0304&51&0.03&Antihistamines, hyposensitisation and allergic emergencies\\
0301&51&0.01&Bronchodilators\\
0302&51&0&Corticosteroids (respiratory)\\
0301&51&0&Bronchodilators\\
0304&51&0&Antihistamines, hyposensitisation and allergic emergencies\\
0302&51&0&Corticosteroids (respiratory)\\
0304&51&0&Antihistamines, hyposensitisation and allergic emergencies\\
0302&51&0&Corticosteroids (respiratory)\\
0301&51&0&Bronchodilators\\
0302&51&0&Corticosteroids (respiratory)\\
0301&51&0&Bronchodilators\\
0302&51&0&Corticosteroids (respiratory)\\
0301&51&0.44&Bronchodilators\\
0501&51&0.2&Antibacterial drugs\\
J441&51&0&Chronic obstructive pulmonary disease with acute exacerbation\\
0301&51&0&Bronchodilators\\
0302&51&0&Corticosteroids (respiratory)\\
0501&51&0.01&Antibacterial drugs\\
J029&51&0&Acute pharyngitis\\
0301&51&0&Bronchodilators\\
0302&51&0&Corticosteroids (respiratory)\\
0301&51&0&Bronchodilators\\
0302&51&0&Corticosteroids (respiratory)\\
0301&52&0&Bronchodilators\\
0302&52&0&Corticosteroids (respiratory)\\
J459&52&0&Asthma\\
0302&52&0&Corticosteroids (respiratory)\\
0301&52&0&Bronchodilators\\
0301&52&0&Bronchodilators\\
0302&52&0.03&Corticosteroids (respiratory)\\
0403&52&0&Antidepressant drugs\\
F329&52&0&Depressive episode\\
0301&52&0&Bronchodilators\\
0302&52&0&Corticosteroids (respiratory)\\
0302&52&0&Corticosteroids (respiratory)\\
0304&52&0&Antihistamines, hyposensitisation and allergic emergencies\\
0301&52&0&Bronchodilators\\
0304&52&0&Antihistamines, hyposensitisation and allergic emergencies\\
0301&52&0&Bronchodilators\\
0302&52&0&Corticosteroids (respiratory)\\
0302&52&0&Corticosteroids (respiratory)\\
0304&52&0.02&Antihistamines, hyposensitisation and allergic emergencies\\
0301&52&0.09&Bronchodilators\\
0301&52&0&Bronchodilators\\
0302&52&0.01&Corticosteroids (respiratory)\\
0302&52&0&Corticosteroids (respiratory)\\
0301&52&0&Bronchodilators\\
0501&52&0&Antibacterial drugs\\
J22&52&0&Acute lower respiratory infection\\
0301&52&0&Bronchodilators\\
0302&52&0.04&Corticosteroids (respiratory)\\
J40&52&0&Bronchitis\\
0302&52&0&Corticosteroids (respiratory)\\
0501&52&0&Antibacterial drugs\\
0301&52&0&Bronchodilators\\
0105/0302/ 0501/0504/ 0603/0802/ 1001&52&0&Chronic bowel disorders/Corticosteroids (respiratory)/ Antibacterial drugs/Antiprotozoal drugs/ Corticosteroids (endocrine)/Drugs affecting the immune response/ Drugs used in rheumatic diseases and gout\\
0501&52&0&Antibacterial drugs\\
0501&52&0&Antibacterial drugs\\
0301&53&0.14&Bronchodilators\\
0302&53&0&Corticosteroids (respiratory)\\
0105/0302/ 0501/0504/ 0603/0802/ 1001&53&0&Chronic bowel disorders/Corticosteroids (respiratory)/ Antibacterial drugs/Antiprotozoal drugs/ Corticosteroids (endocrine)/Drugs affecting the immune response/ Drugs used in rheumatic diseases and gout\\
J441&53&1&Chronic obstructive pulmonary disease with acute exacerbation\\
0501&53&0&Antibacterial drugs\\
0501&53&0&Antibacterial drugs\\
0105/0302/ 0501/0504/ 0603/0802/ 1001&53&0&Chronic bowel disorders/Corticosteroids (respiratory)/ Antibacterial drugs/Antiprotozoal drugs/ Corticosteroids (endocrine)/Drugs affecting the immune response/ Drugs used in rheumatic diseases and gout\\
0501&53&0&Antibacterial drugs\\
0301&53&0.37&Bronchodilators\\
0302&53&0&Corticosteroids (respiratory)\\
J029&53&0.06&Acute bronchitis\\
0501&53&0.01&Antibacterial drugs\\
0301&53&0.01&Bronchodilators\\
0302&53&0&Corticosteroids (respiratory)\\
0304&53&0&Antihistamines, hyposensitisation and allergic emergencies\\
0302&53&0&Corticosteroids (respiratory)\\
0304&53&0&Antihistamines, hyposensitisation and allergic emergencies\\
0301&53&0&Bronchodilators\\
0304&53&0.01&Antihistamines, hyposensitisation and allergic emergencies\\
S526&53&0&Fracture of lower end of both ulna and radius\\
0407&53&0&Analgesics\\
0301&53&0&Bronchodilators\\
0302&53&0.01&Corticosteroids (respiratory)\\
0407/0407&53&0&Analgesics\\
0105/0302/ 0501/0504/ 0603/0802/ 1001&53&0&Chronic bowel disorders/Corticosteroids (respiratory)/ Antibacterial drugs/Antiprotozoal drugs/ Corticosteroids (endocrine)/Drugs affecting the immune response/ Drugs used in rheumatic diseases and gout\\
J441&53&1&Chronic obstructive pulmonary disease with acute exacerbation\\
0501&53&0.01&Antibacterial drugs\\
0501&53&0&Antibacterial drugs\\
J40&53&0&Bronchitis\\
0501&53&0&Antibacterial drugs\\
J029&53&0.07&Acute pharyngitis\\
0501&53&0&Antibacterial drugs\\
0302&53&0.02&Corticosteroids (respiratory)\\
0105/0302/ 0501/0504/ 0603/0802/ 1001&53&0&Chronic bowel disorders/Corticosteroids (respiratory)/ Antibacterial drugs/Antiprotozoal drugs/ Corticosteroids (endocrine)/Drugs affecting the immune response/ Drugs used in rheumatic diseases and gout\\
J40&53&0&Bronchitis\\
0301&53&0&Bronchodilators\\
0501&53&0&Antibacterial drugs\\
0105/0302/ 0501/0504/ 0603/0802/ 1001&53&0&Chronic bowel disorders/Corticosteroids (respiratory)/ Antibacterial drugs/Antiprotozoal drugs /Corticosteroids (endocrine)/Drugs affecting the immune response/ Drugs used in rheumatic diseases and gout\\
0501&53&0&Antibacterial drugs\\
J441&53&0.01&Chronic obstructive pulmonary disease with acute exacerbation\\
J22&54&0.22&Acute lower respiratory infection\\
0501&54&0&Antibacterial drugs\\
0302&54&0&Corticosteroids (respiratory)\\
0301&54&0.41&Bronchodilators\\
J029&54&0&Acute pharyngitis\\
0501&54&0.41&Antibacterial drugs\\
0501/1306&54&0.04&Antibacterial drugs/Acne and rosacea\\
I259&54&0.59&Chronic ischaemic heart disease\\
0202&54&1&Diuretics\\
I219&54&1&Acute myocardial infarction\\
J449&54&1&Chronic obstructive pulmonary disease\\
J459&54&1&Asthma\\
0202&54&1&Diuretics\\
0407/0407&54&0&Analgesics\\
0205&54&0&Anti-hypertensive drugs\\
0301&54&0.02&Bronchodilators\\
0302&54&0&Corticosteroids (respiratory)\\
0301&54&1&Bronchodilators\\
0202&54&0.48&Diuretics\\
0302&54&0&Corticosteroids (respiratory)\\
0301&54&0&Bronchodilators\\
J22&54&0.97&Acute lower respiratory infection\\
0501&54&0&Antibacterial drugs\\
0501&54&0&Antibacterial drugs\\
0202&54&0&Diuretics\\
0301&54&0&Bronchodilators\\

 \bottomrule
\end{supertabular}

\end{document}